\documentclass[final, custom]{anthology-ch} 

\usepackage{booktabs}
\usepackage{graphicx}
\usepackage{booktabs, multirow}
\title{Retrieving Biblical Intertextual References in Karen Blixen’s \textit{Seven Gothic Tales}}

\author[1]{András Kovács}[
  orcid=0009-0007-0931-9535
]

\author[1]{Alexander Conroy}[
  orcid=0000-0001-5693-0571
]

\author[2]{Daniel Hershcovich}[
  orcid=0000-0002-3966-8708
]

\author[1]{Jens Bjerring-Hansen}[
  orcid=0000-0001-5786-8300
]

\affiliation{1}{Department of Nordic Studies and Linguistics, University of Copenhagen, Copenhagen, Denmark}
\affiliation{2}{Department of Computer Science, University of Copenhagen, Copenhagen, Denmark}

\keywords[eng]{computational intertextuality, information retrieval, computational literary studies, biblical intertextuality, Karen Blixen, Danish literature}

\pubyear{2026}
\pagestart{1}
\pageend{1}
\customhead{}  

\begin{document}

\maketitle

\begin{abstract}
Identifying intertextual references is central to literary scholarship, but computationally difficult when source material is transformed through paraphrase, allusion, historical language, and translation. We investigate this problem through biblical intertextuality in Karen Blixen’s \textit{Seven Gothic Tales}. Drawing on the commentary to a critical edition, we construct a benchmark of 189 annotated references and evaluate retrieval against all 31,170 verses of historically plausible Danish Old and New Testament translations. We compare TF--IDF and BM25 with multilingual and Danish sentence encoders, examine the effect of linguistic normalization, and fine-tune a Danish encoder using hard negatives and five-fold cross-validation. We analyze performance across automatically derived lexical-overlap strata representing quotations, paraphrases, and allusions.
Linguistically normalized BM25 provides a strong zero-shot baseline, attaining an overall R@10 of 0.365 and retrieving every quotation within its ten highest-ranked verses. The best zero-shot dense model achieves a comparable overall score of 0.360 while performing better on allusions. Fine-tuning DFM-large raises its overall R@10 from 0.265 to 0.508 and more than doubles its performance on allusions, from 0.138 to 0.339. However, evaluation against editorial annotations alone understates the model’s scholarly usefulness: a literary scholar judged seven of 30 selected rank-one predictions counted as false positives to be meaningful additional references. These findings show both the potential and the epistemic limits of computational intertextual retrieval. Rather than treating scholarly annotations as exhaustive or model outputs as discoveries, we propose retrieval models as heuristic co-readers that recover documented references and generate candidates for expert-led close reading.
\end{abstract}

\section{Introduction}

Literary texts acquire meaning partly through their relationships with earlier texts. These relationships may take the form of direct quotation, paraphrase, or subtler allusion, and recognizing them often depends on historical, cultural, and literary knowledge. Intertextuality therefore shifts attention from an isolated work to the network of texts it absorbs and transforms \cite{kristeva1986word}. Biblical intertextuality is a particularly rich case: biblical language, narratives, and motifs circulate through multiple translations and cultural traditions, frequently without explicit citation. Identifying such references requires what Riffaterre calls “literary competence”---the ability to recognize the source structure underlying an otherwise unusual or semantically marked passage \cite{riffaterre1978semiotics}.

Karen Blixen’s \textit{Seven Gothic Tales} (1935) \cite{blixen2012syv} provides a demanding test case for computational approaches to this task. The tales draw extensively on biblical language and imagery, but the relationship to the Bible ranges from recognizable quotations to highly transformed allusions. Retrieval is further complicated by Blixen’s antiquated Danish, long hypotactic sentences, and the existence of several historically distinct Danish Bible translations. In many cases, the wording familiar to Blixen differs substantially from that of modern translations. Moreover, the relevant source is not always unambiguous: the Bible is highly self-referential, and similar phrases, events, and theological motifs may occur in several books or across both testaments.

Scholarly editions offer a valuable basis for studying these references. The commentary to the critical edition of \textit{Seven Gothic Tales} identifies the biblical passages associated with particular expressions in Blixen’s text. These annotations provide an expert benchmark, but not an exhaustive ground truth. The apparatus was compiled by a single editor, and its purpose is primarily to document established references rather than enumerate every defensible intertextual connection \cite{kondrup2011editionsfilologi}. A computational system may therefore retrieve a passage that differs from the annotated source while remaining relevant to a literary interpretation. This makes biblical intertextuality both an information-retrieval problem and a methodological case for examining how computational similarity relates to scholarly relevance.

In this study, we evaluate whether sparse and dense retrieval methods can recover the biblical references documented in the scholarly edition of \textit{Seven Gothic Tales}. We derive a benchmark of 189 reference pairs and retrieve against all 31,170 verses of the Danish Bible. To provide a historically plausible source text, we use the 1871 translation of the Old Testament and the 1907 translation of the New Testament rather than a contemporary Danish Bible. We compare TF--IDF and BM25 with multilingual and Danish sentence-embedding models, evaluating both raw and linguistically normalized text. We then fine-tune the strongest Danish model using hard negatives and five-fold cross-validation. Performance is analyzed separately for quotations, paraphrases, and allusions, allowing us to examine how the degree of textual transformation affects retrieval.

Our results show that lexical retrieval remains a strong baseline. Preprocessed BM25 obtains an overall Recall@10 of 0.365, marginally outperforming the best zero-shot dense model at 0.360 and recovering every quotation within its ten highest-ranked verses. Fine-tuning substantially improves retrieval, raising overall Recall@10 to 0.508. The improvement is especially pronounced for allusions, where Recall@10 increases from 0.165 for the best zero-shot model to 0.339 after fine-tuning. These results indicate that task-specific training can help models recognize intertextual relationships that extend beyond lexical overlap, even from a comparatively small annotated dataset.

The discrepancies between model predictions and editorial annotations are nevertheless as important as the aggregate scores. Our analysis finds cases in which apparent false positives constitute plausible alternative references, including passages that express the same biblical event or connect an ostensibly New Testament reference to an earlier Old Testament source. A retriever that is agnostic to established theological citation conventions may consequently obscure some culturally expected relationships while revealing others. We therefore argue that computational retrieval should not be understood simply as automating editorial annotation. Its more productive role is that of a heuristic co-reader: recovering established references, proposing alternatives, and making the assumptions behind different notions of intertextual relevance available for scholarly examination.

\section{Related Work}

Computational approaches to intertextuality have developed from overlap-based detection towards vector representations capable of modelling semantic similarity. Early systems such as Tesserae \cite{coffee_tesserae_2013} identify candidate parallels through shared words and related lexical features. Such matches are readily interpretable, since scholars can inspect the textual evidence underlying each result. TRACER similarly supports text-reuse detection using configurable similarity measures. Miyagawa et al. \cite{miyagawa_exploring_2024} applied TRACER alongside Word2Vec and stylometric analysis to Vedic Sanskrit literature, demonstrating the value of combining complementary forms of evidence. Lexical methods are particularly effective for quotations and near-verbatim reuse, but their reliance on observable overlap makes them less suitable for detecting synonymy, paraphrase, and allusion \cite{duan_quantitative_2025}.

Linguistic normalisation can extend the reach of both lexical and representation-based methods, especially for historical languages characterised by substantial morphological and orthographic variation. Burns et al. \cite{burns_profiling_2021} trained an optimised Word2Vec model on lemmatised Latin, achieving state-of-the-art performance on synonym detection and outperforming a widely used lexical method for intertextual search. Their work illustrates how distributional representations can recover relationships beyond exact word overlap. However, static word embeddings do not model words in context, and word-level similarities must still be aggregated into a score for the passage as a whole.

More recent work has therefore turned to contextual encoders and task-specific fine-tuning. D'Angelo et al. \cite{taddei_detecting_nodate} fine-tuned a contextual encoder for semantic reuse in Ancient Greek using contrastive learning. Their training data combined automatically generated paraphrases with hard negatives: passages with substantial lexical similarity but different meanings. This teaches the model to distinguish semantic equivalence from superficial overlap and demonstrates how task-specific representations can be learned even where manually annotated data are scarce. More generally, Duan \cite{duan_quantitative_2025} surveys hybrid approaches that combine lexical, sequential, and semantic similarity measures, suggesting that sparse and dense representations offer complementary evidence rather than forming a simple progression in which one replaces the other.

Generative LLMs can extend this process from retrieving possible references to interpreting their significance. Umphrey et al. \cite{umphrey_investigating_2024} used an expert-in-the-loop methodology to identify and analyse quotations, allusions, and echoes within Koine Greek biblical texts. The models produced novel intertextual observations, but also struggled with long query passages and proposed false dependencies, underscoring the continued need for expert evaluation. Werner and Reiter \cite{werner_between_2025} similarly combined semantic-similarity retrieval with prompt-based analysis of possible relationships between Virginia Woolf's \textit{Mrs Dalloway} and Homer's \textit{Odyssey}. Supplying literary theory and documented examples increased the model's sensitivity to subtle relationships and improved the theoretical grounding of its analyses. At the same time, the expert-informed prompt showed a greater tendency towards over-interpretation, illustrating how scholarly framing can both guide and bias model judgements.

Taken together, previous work shows that lexical retrieval offers precision and interpretability, dense representations broaden retrieval to less explicit forms of reuse, and generative models can support their subsequent interpretation. However, these approaches have rarely been compared systematically on a single expert-annotated literary benchmark that distinguishes quotations, paraphrases, and allusions. Moreover, relatively little work has examined intertextual retrieval where both the literary language and the relevant source tradition vary historically. We address these gaps through a controlled comparison of lexical and dense retrieval methods for biblical references in Karen Blixen's \textit{Syv fantastiske Fortællinger}, examining the effects of linguistic preprocessing and task-specific fine-tuning across different forms of intertextuality.

\section{Source Texts and Benchmark Construction}

The data for this study come from the scholarly edition of Karen Blixen’s \textit{Syv fantastiske Fortællinger} (1935; first published in English as \textit{Seven Gothic Tales} in 1934) \cite{blixen2012syv}. This is the third volume in the series \textit{Karen Blixen. Værker} (2007--2020), established by Det Danske Sprog- og Litteraturselskab (the Society for Danish Language and Literature). The edition is based on the first Danish edition, published in 1935, and includes limited emendations, primarily supported by the manuscript of the book and Blixen’s own copy, both of which contain handwritten corrections by the author. It preserves the original orthography and punctuation. Consequently, the models are exposed to complex grammatical and syntactic structures, as Blixen deliberately employed an antiquated style characterized by long, hypotactic sentences.

Importantly, the commentary apparatus, in which the biblical references are identified, was peer reviewed. However, the apparatus is the work of a single editor, meaning that responsibility for identifying the intertextual references was not shared \cite{kondrup2011editionsfilologi}. Although the edition therefore represents the most thorough search for and documentation of Blixen’s biblical references to date, some references may nevertheless have been overlooked. The annotations provide the benchmark against which we evaluate the models, but they should not be understood as a definitive account of all possible biblical references in the text.

Following the conventions of German and Scandinavian scholarly editing, the notes in \textit{Karen Blixen. Værker} keep interpretation to a minimum. Notes that highlight intertextuality merely locate and describe the source passage and quote the relevant biblical text. In some cases, these descriptions are accompanied by references to other culturally significant engagements with the same biblical passages, such as those of the Danish philosopher Søren Kierkegaard. The edition also identifies overlaps within Blixen’s own authorship when references to the same biblical passages recur across her works. This adds an intra-authorial dimension to the intertextual network by connecting passages throughout Blixen’s body of work. Both forms of reference underscore that intertextuality---and biblical intertextuality in particular---is not a straightforward relationship between a source text and an intertext, but rather a mosaic of textual patterns formed across time and place \cite{kristeva1986word}. Questions concerning how or why Blixen makes these intertextual connections, whether biblical or otherwise literary, are deliberately left unanswered in the notes and instead left to readers’ interpretation.

From a technical perspective, three characteristics of the data complicate the identification of biblical references in Blixen’s gothic tales. These primarily concern the degree of semantic similarity between Blixen’s text and the Bible. First, some intertextual references draw on verses from multiple locations in the Bible. In \textit{Seven Gothic Tales}, the editor identifies 23 cases in which an intertext refers to two or three distinct biblical verses. To avoid making these cases inherently two or three times more difficult than annotations with a single target, we split them according to the number of biblical references, allowing each reference to be evaluated independently. After this procedure, the benchmark contains 189 Blixen--Bible reference pairs.

Second, the length of the annotated passages from Blixen varies considerably (see Figure~\ref{fig:reference_length}). Across all instances, the median length is seven words, but the passages range from a single word---for example, ‘Morderengel’ (‘murder angel’)---to multiple sentences, with the longest annotated example extending to 204 words.

Third, and most importantly, there is the question of translation: specifically, which Danish Bible translation should serve as the source text for probing intertextuality. \textit{Karen Blixen. Værker} primarily refers to and quotes from the 1992 Danish Bible translation, which was the most recent translation available at the time of publication and therefore the most accessible to its readers. However, the notes also demonstrate that semantic similarity is in many cases obscured by newer translations of the Bible, a problem the edition attempts to address by referring readers to older translations. For \textit{Seven Gothic Tales}, these include the New Testament translations of 1819 and 1907 and the Old Testament translations of 1740, 1871, and 1931. In addition to representing diachronically diverse forms of Danish, these Bible editions reflect different paradigms of biblical translation, with varying assumptions regarding the balance between linguistic form and semantic content, particularly the extent to which a translation should be verbatim or idiomatic.

Moreover, we do not know which Bible translation Blixen herself preferred; whether she primarily engaged with the Bible in Danish or English---\textit{Seven Gothic Tales} was, after all, first written in English---or the extent to which her biblical references arose through direct reading of the Bible rather than second-hand accounts or cultural memory. Despite these uncertainties, and in contrast to \textit{Karen Blixen. Værker}, we use the 1871 Old Testament and the 1907 New Testament as our source texts. Both were in circulation during the period in which Blixen wrote her tales and thus provide a more historically plausible point of comparison than the 1992 translation. Together, they form a retrieval corpus of 31,170 verses. This choice does not imply that these editions were necessarily Blixen’s sources; rather, it provides a consistent and historically plausible corpus against which to evaluate retrieval.

The importance of using historically appropriate translations is illustrated by a 2026 trial translation of the new complete Danish Bible, planned for publication in 2036. The hymn ‘Hil dig, Frelser og Forsoner’ (‘Hail You, Saviour and Atoner’) by the nineteenth-century Danish poet N.F.S. Grundtvig echoes an older phrasing of the soldiers’ mocking salutation to Jesus: ‘Hil dig, jødekonge’ (‘Hail, King of the Jews’). In the proposed translation, however, the passage is rendered as ‘Længe leve jødernes konge’ (‘Long live the King of the Jews’). When the hymn is compared with this modern rendering, the lexical cue disappears completely, making the intertextual relationship invisible to human readers as well as an overlap-based retrieval system.

\section{Operationalizing Biblical Intertextuality}

Intertextuality shifts the analytical focus from the isolated text to the network of relationships between texts. Introduced by Julia Kristeva, the concept suggests that any text is constructed as a ``mosaic of quotations,'' absorbing and transforming other discourses \cite{kristeva1986word}. In literary studies, this framework decenters the author by treating meaning-making as a dynamic process that depends on relational networks rather than isolated invention.

To operationalize this broad concept for computational analysis, more differentiated structural accounts are needed to categorize specific forms of textual linking. Gérard Genette's framework of ``transtextuality'' provides a taxonomy of structural relationships and traceable derivations between texts \cite{genette1997palimpsests}. Fairclough further distinguishes between ``manifest intertextuality,'' in which specific texts are explicitly incorporated through mechanisms such as quotation, and ``interdiscursivity,'' which involves the incorporation of broader conventions, genres, or discourses \cite{fairclough1992intertextuality}.

These frameworks do not provide a ready-made annotation scheme for computational retrieval. We therefore adapt, rather than directly reproduce, their distinctions in our taxonomy of biblical references in Karen Blixen's texts. We distinguish three categories. A \emph{quotation} reproduces distinctive wording from a biblical source passage, allowing for minor linguistic variation. A \emph{paraphrase} reformulates the content of a localized source passage without preserving its wording. An \emph{allusion} evokes a particular biblical passage, event, figure, or motif more indirectly. Quotations represent the clearest instances of manifest intertextuality, while paraphrases and allusions involve progressively less explicit relationships to an identifiable source. We exclude broader, work-level strategies such as parody and pastiche, as well as questions of plagiarism. However, localized textual overlaps contributing to such broader relationships may still be captured by our three categories.

Operationalizing these distinctions is inherently difficult because intertextuality poses significant classification challenges even for human readers. The boundaries between categories---for example, between an explicit paraphrase and an implicit allusion---are often blurred, making fully consistent annotation elusive. Because identifying these relationships depends on interpretation, Michael Riffaterre locates intertextuality in the reader's encounter with the text. He describes it as a reading mechanism triggered by semantic anomalies, or ``ungrammaticalities,'' whose resolution requires the reader to possess the ``literary competence'' needed to identify an underlying source structure, or hypogram \cite{riffaterre1978semiotics}. Replicating this implicit and culturally situated competence in NLP models remains a central challenge.

These theoretical and technical difficulties are especially pronounced in \textit{Syv fantastiske Fortællinger}. The annotated passages vary substantially in length, ranging from a single word to spans exceeding 200 words. Similarity between a Blixen passage and its biblical source may also be obscured by Blixen's deliberately antiquated style, including complex, hypotactic sentences, as well as by the historical and translational distance between different Danish versions of the Bible. Computational models must therefore identify relationships that may be lexical, semantic, narrative, or culturally mediated across passages of markedly different lengths and linguistic forms.

Accordingly, our computational task is deliberately narrower than intertextuality as a general literary-theoretical concept. Given an annotated passage from Blixen, the models rank candidate verses from the Danish Bible, and we evaluate whether they retrieve the source passage identified in the scholarly commentary. Failure to retrieve that passage therefore indicates disagreement with the benchmark rather than the absence of any plausible intertextual relationship. Conversely, a highly ranked but unannotated verse is not necessarily irrelevant. This distinction is particularly important because the commentary provides an expert benchmark, but not a definitive account of every possible biblical reference in Blixen's work.

\section{Reference Distribution and Automatic Classification}

\subsection{Reference Counts and Distribution}

The nearly 250 pages of editorial commentary contain numerous annotations. Using automatic extraction based on regular-expression patterns for biblical citations, we identified 164 annotations containing references to the Bible. Three refer to the biblical apocrypha rather than either the Old or New Testament and were therefore excluded, leaving 161 annotations.

Some annotations identify several distinct, non-contiguous locations in the Bible. For example, ``den Orm, som aldrig dør'' (``the worm that never dies'') is linked to passages in both Isaiah in the Old Testament and the Gospel of Mark in the New Testament. Because each retrieval instance requires a single target passage, we split such annotations by biblical location. In total, 23 annotations contained multiple separate targets, some of them more than two. Splitting these annotations produced 28 additional instances, bringing the benchmark from 161 annotations to 189 reference instances.

A \emph{reference instance} is thus defined as a pairing between one annotated Blixen lemma and one continuous biblical passage. The 189 instances represent annotated relationships rather than 189 distinct biblical passages or verses: instances are counted separately when the same verse is referenced more than once. Most individual verses occur in only one reference instance. The most frequently referenced verse, Genesis 1:1 (1~Mosebog 1:1), occurs in 13 instances.

Of the 189 reference instances, 106 target a single Bible verse, while 83 target continuous passages containing between two and ten verses. Among the 83 multi-verse references, 43 span exactly two verses, and most span no more than six. There is one instance each spanning seven, eight, and ten verses.

As mentioned above, most of the annotated Blixen lemmas are relatively short. They range from 1 to 204 words, with a median length of 7 words. At the extremes, 10 consist of a single word, while 13 contain more than 40 words. The corresponding biblical passages also vary considerably in length. They range from 5 to 194 words, with a median length of 34 words, making them considerably longer on average than the Blixen lemmas. The median is 26 words for single-verse references and 57 words for multi-verse references.

Both sides of this reference network are themselves collections: \textit{Seven Gothic Tales} consists of seven stories, while the Bible corpus contains 66 books. Examining the distribution across these works shows that \textit{Syndfloden over Norderney}, in which one of the main characters is a cardinal, contains the largest number of reference instances, with 63. Among the biblical books, the Gospel of Matthew is referenced most frequently, with 71 instances, followed by Genesis (the First Book of Moses), with 26. In total, the annotations refer to 24 of the 66 books in the biblical canon.

\begin{figure}
\centering
\includegraphics[width=\linewidth]{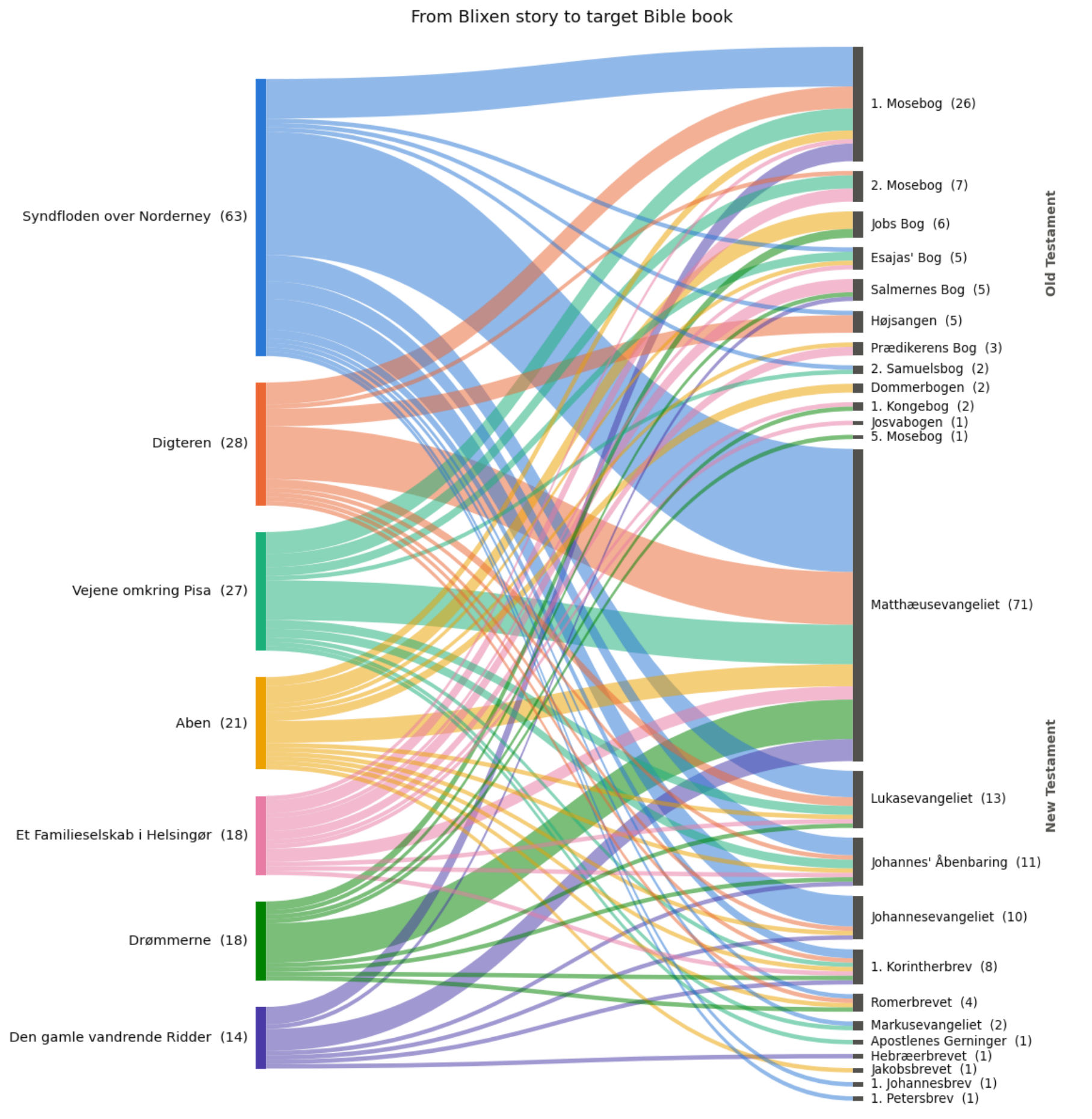}
\caption{Sankey diagram of the 189 reference instances linking stories in \textit{Seven Gothic Tales} to books of the Bible. Both the stories and the biblical books are sorted in descending order by their number of references. The biblical books are additionally grouped by testament. The displayed value in parentheses is the number of reference instances. Each instance is counted separately even when the same biblical verse is referenced multiple times; consequently, the flows sum to 189 on both sides of the diagram.}
\label{fig:bible-book}
\end{figure}

Grouped by testament, 65 reference instances target the Old Testament, comprising 40 single-verse and 25 multi-verse references. The remaining 124 target the New Testament, comprising 66 single-verse and 58 multi-verse references. These counts sum to the complete set of 189 reference instances and are illustrated in Figure~\ref{fig:bible-book}.

\subsection{Automatic Classification of Relationship Types}\label{sec:knowledge}

As discussed above, we distinguish between quotations, paraphrases, and allusions. To assign these relationship types consistently across the 189 reference instances, we constructed an automatic classification pipeline based on lexical overlap. For each pairing of a Blixen lemma and a biblical passage, we calculated the Jaccard index between their processed token sets.

Both texts were first lemmatized using the small DaCy model \cite{f975f4ce65944e3ea958578003cee622} and then stemmed using the Danish Snowball stemmer. This normalization was intended to reduce inflectional and diachronic variation between the two works. If $A$ and $B$ denote the sets of unique processed tokens in the Blixen lemma and the biblical passage, respectively, their Jaccard similarity is
\[
J(A,B)=\frac{|A\cap B|}{|A\cup B|}.
\]

The score ranges from 0, when the texts share no processed tokens, to 1, when their token sets are identical.

We then applied $k$-means clustering with $k=3$ to the one-dimensional distribution of Jaccard scores. The resulting cluster boundaries were approximately 0.093 and 0.296. For the subsequent analyses, we rounded these values to fixed thresholds of 0.1 and 0.3. Instances with a score below 0.1 were classified as allusions, those with a score from 0.1 up to 0.3 as paraphrases, and those with a score of at least 0.3 as quotations. This procedure produced 109 allusions, 54 paraphrases, and 26 quotations.

This classification has important limitations. In particular, the Blixen lemmas are typically shorter than the biblical passages. The Jaccard index therefore penalizes references to longer passages: even when every unique word in a Blixen lemma occurs in the corresponding biblical passage, the score is limited by the additional words in that passage. Formally, when $A\subseteq B$, the score is simply $|A|/|B|$. Lemmatization and stemming may also collapse meaningful linguistic distinctions, while lexical overlap alone cannot establish whether a relationship is properly understood as a quotation, paraphrase, or allusion.

The resulting labels should therefore be understood as reproducible, automatically derived approximations of the theoretical taxonomy---or, more precisely, as lexical-overlap strata---rather than as expert-validated literary classifications. They serve as an analytical tool for comparing retrieval performance across different degrees of lexical explicitness, but not as a definitive interpretation of each intertextual relationship.

\section{Experimental Setup}

We conducted two sets of experiments. First, we evaluated sparse and dense retrieval methods in a zero-shot setting, without training on the Blixen benchmark. Second, we fine-tuned the strongest Danish sentence encoder to assess the effect of task-specific training on intertextual retrieval.

For each of the 189 reference instances, the annotated passage from Blixen---hereafter the \emph{query}---was used to rank all 31,170 verses in the Bible corpus. Sparse methods ranked verses primarily according to lexical overlap, whereas dense methods ranked them according to similarity between contextual embeddings. The gold-standard target for each instance was the continuous biblical passage identified in the editorial commentary. For multi-verse targets, every verse in the annotated passage was treated as relevant, and retrieval was considered successful if at least one of these verses appeared among the retrieved results.

\subsection{Preprocessing}

We applied orthographic normalization to both the Blixen queries and the Bible corpus to reduce mismatches between historical Danish and the predominantly modern Danish on which the models were trained. Most notably, the historical digraph \textit{aa} was converted to the modern letter \textit{\aa}. At least one such conversion occurred in 57\% of the Bible verses.

For sparse retrieval, we additionally compared two text representations. The surface-form representation retained the words after orthographic normalization, while the linguistically normalized representation was lemmatized using the small DaCy model and subsequently stemmed using the Danish Snowball stemmer. This was the same linguistic-normalization pipeline used to calculate the Jaccard similarities described above. Dense models received the orthographically normalized surface text rather than the lemmatized and stemmed representation.

\subsection{Sparse Retrieval}

We evaluated TF--IDF and BM25, two established bag-of-words retrieval methods \cite{salton1988term,robertson_probabilistic_2009}. TF--IDF weights terms according to their frequency within a verse and their inverse frequency across the corpus. BM25 additionally applies term-frequency saturation and explicitly normalizes scores according to document length. This reduces the tendency to favour longer verses merely because they contain more query terms. We evaluated both methods using the surface-form and linguistically normalized text representations described above.

\subsection{Dense Retrieval}

For zero-shot dense retrieval, we evaluated small and large variants from two model families: Multilingual E5 \cite{wang2024multilingual} and the Danish Foundation Models (DFM) \cite{enevoldsen2023danish} sentence encoders. The Multilingual E5 models are initialized from the corresponding XLM-RoBERTa architectures. The small and large variants contain approximately 118 and 560 million parameters, respectively. They were contrastively pretrained on one billion multilingual text pairs and subsequently fine-tuned on supervised retrieval and semantic-similarity datasets \cite{wang2024multilingual}.

The DFM-small sentence encoder contains approximately 22 million parameters and is based on a Danish ELECTRA model. DFM-large contains approximately 355 million parameters and derives from NB-BERT-large through continued pretraining on the Danish Colossal Corpus and subsequent adaptation for sentence embeddings. The sentence-embedding adaptation of DFM-large used SimCSE training on paragraphs extracted from the Danish Gigaword Corpus \cite{enevoldsen2023danish,derczynski-etal-2021-danish}. For each model, the Blixen queries and Bible verses were encoded independently, and the verses were ranked by cosine similarity between their embeddings.

The available documentation does not identify the models' pretraining data at the level of individual literary or biblical works. We therefore cannot determine whether the precise Bible translations or passages from Blixen used in this study occurred in their pretraining corpora. Although Blixen's works remain under copyright in Denmark until the end of 2032, their copyright status alone cannot establish their absence from model-training data. The zero-shot evaluation should consequently not be interpreted as providing a guarantee against pretraining overlap.

\subsection{Fine-Tuning}

We fine-tuned DFM-large as a bi-encoder using multiple-negatives ranking loss and mined hard negatives \cite{reimers-gurevych-2019-sentence}. The objective brings each Blixen query closer to its gold-standard biblical passage while separating it from hard-negative verses and the positive passages associated with other queries in the same batch.

Because the benchmark contains only 189 reference instances, we used five-fold cross-validation. The folds were grouped by the original editorial annotation, ensuring that reference instances produced by splitting a single annotation with several non-contiguous biblical targets were assigned to the same fold. The folds were also stratified, as far as the grouping constraint permitted, by the automatically assigned relationship type---quotation, paraphrase, or allusion---to maintain approximately comparable distributions across folds.

For each fold, the model was trained on approximately $\frac{4}{5}$ of the reference instances. Each held-out query was then evaluated by ranking the complete corpus of 31,170 Bible verses. We concatenated the predictions from the five held-out folds before calculating the aggregate metrics. Thus, every reference instance was evaluated using a model that had not been trained on its source annotation.

Grouping was performed by editorial annotation rather than by biblical verse. Because the same verse may be referenced in several independent annotations, a verse used as a positive example in one training fold may also be relevant to a query in the corresponding test fold. The experiment therefore evaluates generalization to unseen annotated passages from Blixen, rather than to previously unseen Bible verses.

\subsection{Evaluation Metrics}

Let $G_q$ denote the set of gold-standard verses for query $q$, and let $r_i(q)$ denote the verse retrieved at rank $i$. We define binary relevance as
\[
\operatorname{rel}_i(q)=
\begin{cases}
1, & \text{if } r_i(q)\in G_q,\\
0, & \text{otherwise}.
\end{cases}
\]

We report the following metrics, macro-averaged over the 189 reference instances:

\begin{itemize}
\item \textbf{Precision at 1 (P@1):} whether the highest-ranked verse belongs to $G_q$. Because only one result is considered, this is numerically equivalent to Hit@1.

\item \textbf{Recall at 10 (R@10):} a binary, instance-level measure indicating whether at least one verse from \(G_q\) occurs among the ten highest-ranked verses. More precisely, this is a Hit@10 measure rather than conventional set recall, since retrieving additional verses from a multi-verse target does not increase the score.

\item \textbf{Mean reciprocal rank at 10 (MRR@10):} the reciprocal rank of the first retrieved verse belonging to \(G_q\), or zero if no relevant verse occurs among the first ten results.

\item \textbf{Normalized discounted cumulative gain at 10 (nDCG@10):} a rank-sensitive measure that gives credit for every relevant verse retrieved within the first ten positions \cite{jarvelin2002cumulated}. We use binary relevance and normalize against an ideal ranking in which up to \(\min(|G_q|,10)\) gold-standard verses occupy the highest positions.

\end{itemize}

The distinction between R@10 and nDCG@10 is particularly relevant for multi-verse targets: R@10 records whether the target passage was found at all, whereas nDCG@10 also reflects how many of its constituent verses were retrieved and how highly they were ranked.

\section{Results}

We first compare the zero-shot retrieval methods using R@10, which indicates whether the gold-standard biblical passage is represented among the ten highest-ranked verses. We then compare the best zero-shot systems with the fine-tuned model across all evaluation metrics. Finally, we report an expert reassessment of a selected set of apparent false-positive predictions. The quotation, paraphrase, and allusion categories used below are the automatically derived relationship types described in Section~\ref{sec:knowledge}.

\subsection{Zero-Shot Retrieval}

\subsubsection{Sparse Retrieval}

Table~\ref{tab:sparse} compares TF--IDF and BM25 with surface-form text and with linguistic normalization through lemmatization and stemming. Normalization improved overall R@10 for both methods: from 0.286 to 0.302 for TF--IDF and from 0.286 to 0.365 for BM25. However, the effect was not uniform across relationship types. In particular, TF--IDF performance on paraphrases decreased from 0.500 to 0.426.

BM25 with lemmatized and stemmed text obtained the strongest overall sparse-retrieval result, with a R@10 of 0.365. It retrieved at least one relevant verse within the first ten results for all 26 quotations and performed better than the other sparse configurations on both paraphrases and allusions. We therefore use this configuration as the sparse baseline in the subsequent comparison.

\begin{table}[t]
\centering
\small
\begin{tabular}{lcccc}
\toprule
\textbf{Model} & \textbf{Quotation} & \textbf{Paraphrase} & \textbf{Allusion} & \textbf{All} \\
\midrule
TF--IDF raw      & 0.769 & 0.500 & 0.064 & 0.286 \\
TF--IDF lem.+stem & 0.923 & 0.426 & 0.092 & 0.302 \\
BM25 raw         & 0.692 & 0.537 & 0.064 & 0.286 \\
BM25 lem.+stem   & \textbf{1.000} & \textbf{0.574} & \textbf{0.110} & \textbf{0.365} \\
\bottomrule
\end{tabular}
\caption{Sparse-retrieval R@10 by automatically assigned relationship type (quotation (n=26), paraphrase (n=54), allusion (n=109), and all (n=189)).}
\label{tab:sparse}
\end{table}

\subsubsection{Dense Retrieval}

Table~\ref{tab:dense} presents the zero-shot dense-retrieval results. Within both model families, the large variant outperformed the corresponding small variant across all relationship types. This comparison does not, however, isolate the effect of parameter count, since the variants may also differ in their architectures and training data.

The strongest dense model was e5-large, with an overall R@10 of 0.360. It substantially outperformed dfm-large, which obtained 0.265, and achieved the best zero-shot result on allusions, at 0.165. Nevertheless, its overall result remained slightly below that of the linguistically normalized BM25 system, at 0.360 compared with 0.365. Thus, the best sparse and dense zero-shot systems performed similarly overall but exhibited different strengths: BM25 was stronger on quotations and paraphrases, whereas e5-large performed better on allusions.

\begin{table}[t]
\centering
\small
\begin{tabular}{lcccc}
\toprule
\textbf{Model} & \textbf{Quotation} & \textbf{Paraphrase} & \textbf{Allusion} & \textbf{All} \\
\midrule
e5-small  & 0.538 & 0.222 & 0.119 & 0.206 \\
e5-large  & \textbf{0.808} & \textbf{0.537} & \textbf{0.165} & \textbf{0.360} \\
dfm-small & 0.192 & 0.074 & 0.028 & 0.063 \\
dfm-large & 0.692 & 0.315 & 0.138 & 0.265 \\
\bottomrule
\end{tabular}
\caption{Zero-shot dense-retrieval R@10 by automatically assigned relationship type. Sample sizes are identical to those in Table~\ref{tab:sparse}.}
\label{tab:dense}
\end{table}

\subsection{Effect of Fine-Tuning}

Table~\ref{tab:main} compares the strongest sparse and dense zero-shot systems with dfm-large before and after fine-tuning. Fine-tuning improved dfm-large across every metric and relationship type. Its overall R@10 increased from 0.265 to 0.508, corresponding to an increase from 50 to 96 of the 189 reference instances. Its overall P@1 increased from 0.116 to 0.339, meaning that the gold-standard passage contained the highest-ranked verse for 64 instances.

The largest improvement was observed for the less lexically explicit relationships. For paraphrases, fine-tuning increased R@10 from 0.315 to 0.685. For allusions, it increased R@10 from 0.138 to 0.339. The latter result is slightly more than twice the best zero-shot dense result of 0.165 obtained by e5-large.

The fine-tuned model achieved the strongest overall result for every metric and the strongest results on paraphrases and allusions. BM25 nevertheless remained stronger on quotations for R@10, MRR@10, and nDCG@10, while the two systems obtained the same quotation P@1 of 0.654. These results suggest that lexical retrieval remains particularly effective for explicit quotations, whereas task-specific fine-tuning is most beneficial for paraphrases and allusions.

\begin{table}[t]
\centering
\small
\begin{tabular}{llcccc}
\toprule
\textbf{Model} & \textbf{Metric} & \textbf{Quotation} & \textbf{Paraphrase} & \textbf{Allusion} & \textbf{All} \\
\midrule
\multirow{4}{*}{BM25 lem.+stem}
& P@1     & \textbf{0.654} & 0.241 & 0.037 & 0.180 \\
& R@10  & \textbf{1.000} & 0.574 & 0.110 & 0.365 \\
& MRR@10  & \textbf{0.811} & 0.337 & 0.056 & 0.240 \\
& nDCG@10 & \textbf{0.860} & 0.393 & 0.068 & 0.270 \\
\midrule
\multirow{4}{*}{e5-large}
& P@1     & 0.462 & 0.148 & 0.055 & 0.138 \\
& R@10  & 0.808 & 0.537 & 0.165 & 0.360 \\
& MRR@10  & 0.562 & 0.262 & 0.095 & 0.207 \\
& nDCG@10 & 0.620 & 0.327 & 0.112 & 0.243 \\
\midrule
\multirow{4}{*}{dfm-large}
& P@1     & 0.385 & 0.148 & 0.037 & 0.116 \\
& R@10  & 0.692 & 0.315 & 0.138 & 0.265 \\
& MRR@10  & 0.487 & 0.189 & 0.064 & 0.158 \\
& nDCG@10 & 0.537 & 0.218 & 0.082 & 0.183 \\
\midrule
\multirow{4}{*}{dfm-ft}
& P@1     & \textbf{0.654} & \textbf{0.481} & \textbf{0.193} & \textbf{0.339} \\
& R@10  & 0.846 & \textbf{0.685} & \textbf{0.339} & \textbf{0.508} \\
& MRR@10  & 0.737 & \textbf{0.549} & \textbf{0.229} & \textbf{0.390} \\
& nDCG@10 & 0.765 & \textbf{0.581} & \textbf{0.255} & \textbf{0.418} \\
\bottomrule
\end{tabular}
\caption{Comparison of BM25 with lemmatization and stemming, e5-large, and dfm-large before and after fine-tuning. Bold indicates the highest score within each metric and relationship type; ties are both highlighted. The dfm-ft results are out-of-fold predictions from five-fold cross-validation grouped by editorial annotation.}
\label{tab:main}
\end{table}

\subsection{Expert Reassessment of Apparent False Positives}

The primary evaluation assumes that only the verses identified in the scholarly commentary are relevant. As discussed above, however, the commentary is not necessarily an exhaustive account of every biblical reference in Blixen's work. We therefore examined whether some apparent retrieval errors represented plausible additional references.

A literary scholar reassessed 30 cases selected from the apparent P@1 errors made by dfm-ft. In every case, the verse ranked first by the model was absent from the original gold-standard passage and was therefore treated as non-relevant in the primary evaluation. The expert reconsidered each retrieved verse in relation to the corresponding Blixen passage and assigned a binary relevance judgement. The selected cases comprised seven rows classified as paraphrases and 23 classified as allusions; no quotation cases were included.

The expert judged 7 of the 30 top-ranked verses to be relevant, resulting in an expert-reassessed P@1 of 0.233 on this selected subset. This included 2 of the 7 paraphrase cases and 5 of the 23 allusion cases, corresponding to P@1 values of 0.286 and 0.217, respectively.

\begin{table}[t]
\centering
\small
\begin{tabular}{llrrr}
\toprule
\textbf{Model} & \textbf{Relationship type} & \(\boldsymbol{n}\) & \textbf{Relevant} & \textbf{P@1} \\
\midrule
\multirow{3}{*}{dfm-ft}
& Paraphrase & 7  & 2 & 0.286 \\
& Allusion   & 23 & 5 & 0.217 \\
& All        & 30 & 7 & 0.233 \\
\bottomrule
\end{tabular}
\caption{Expert-reassessed P@1 for 30 selected dfm-ft predictions that were counted as incorrect under the original benchmark. ``Relevant'' gives the number of top-ranked verses subsequently judged relevant by the literary scholar.}
\label{tab:expert-sample}
\end{table}

These findings show that some predictions counted as false positives under the original benchmark were considered plausible references upon expert reassessment. However, the 30 cases were selected from dfm-ft's apparent rank-one errors rather than sampled independently from the complete benchmark. The resulting P@1 values are therefore descriptive of this selected subset and should not be interpreted as an unbiased estimate of performance on the full dataset. Because only dfm-ft's highest-ranked candidates were reassessed, the analysis also cannot support comparisons with the other retrieval methods or an adjusted full-benchmark score.

\section{Analysis and Discussion}

\subsection{Effects of Fine-Tuning}

Fine-tuning substantially improved retrieval, particularly for paraphrases and allusions, for which lexical overlap provides a weak signal. Individual examples illustrate the scale of this improvement. The Blixen lemma ``Behemoth og Leviatan'' refers to the ten-verse passage Job 40:10--19. The zero-shot dfm-large model ranked the first verse of this ground-truth passage at position 430, whereas dfm-ft ranked a verse from the passage first. Fine-tuning can therefore enable the model to retrieve relevant passages despite substantial differences in wording and passage length.

This example does not, however, establish that fine-tuning specifically resolved length discrepancies or the structural ``ungrammaticalities'' discussed above. The retrieval results reveal which passages the model ranks, but not which textual features caused the improvement. As discussed below, the aggregate analysis also provides no indication that dfm-ft performs systematically better on shorter or longer passages. The result is therefore better interpreted as evidence that task-specific training improved retrieval across heterogeneous references, rather than as evidence for a particular learned mechanism.

\subsection{The Selectivity of the Ground Truth}

The primary evaluation operationalizes the ground truth as the biblical passage identified in the expert commentary. This provides a reproducible benchmark, but it does not imply that the identified passage is the only plausible intertext. Because the commentary reflects the judgements and aims of a particular scholarly edition, it is necessarily selective. A retrieved verse absent from the benchmark may consequently be either an irrelevant match or a plausible additional reference.

The expert reassessment reported in the preceding section supports this distinction. Among the 30 selected dfm-ft predictions counted as P@1 errors under the original benchmark, seven were subsequently judged relevant by a literary scholar. These comprised two of the seven paraphrase cases and five of the 23 allusion cases. The finding does not provide an adjusted estimate of performance, because the examples were deliberately selected from dfm-ft's apparent errors. It nevertheless demonstrates that the benchmark produces some false negatives: predictions counted as incorrect may identify meaningful relationships not recorded in the original commentary. Table~\ref{tab:expert-sample} presents examples of these additional matches.

For example, the commentary links the Blixen lemma `staaet op fra de Døde'' (`risen from the dead'') to Matthew 28:1--7. Dfm-ft instead ranked Acts 13:30---`Men Gud oprejste ham fra de døde'' (`But God raised him from the dead'')---first. The literary scholar judged this prediction relevant because both passages refer explicitly to Christ's resurrection. This does not make the original annotation incorrect: Matthew remains the annotated source, while Acts constitutes an additional relevant parallel. The example instead illustrates how a retrieval model can serve as a heuristic co-reader by proposing candidates for further scholarly examination.

\subsection{Biblical Self-Reference and Interpretive Conventions}

Alternative matches are particularly common because the Bible is highly self-referential. Figures, formulations, and events recur across books, while New Testament passages frequently reinterpret or present themselves as fulfilling material from the Old Testament. A retrieval model may therefore identify a closely related passage without reproducing the particular source attribution preferred by the expert commentary.

The reference to ``the new Jerusalem'' illustrates this problem. The ground-truth annotation identifies the passage in Revelation, whereas the model retrieves an earlier related passage from Isaiah. Relative to the benchmark, the prediction is an error. From an exploratory intertextual perspective, however, Isaiah may still be relevant to the formulation and its textual history. The model appears to prioritize lexical and semantic similarity without representing the theological convention by which an Old Testament prophecy may be cited through its presumed New Testament fulfilment.

Whether this behaviour is desirable depends on the research question. If the objective is to reproduce the source attributions in the scholarly commentary, alternative biblical parallels should be treated as errors. If the objective is to discover possible intertexts, retrieving an earlier or less conventional parallel may be analytically valuable. This distinction is especially pertinent when investigating whether Blixen draws on Old Testament material in contexts where Christian interpretive convention might instead privilege a New Testament formulation. Semantic retrieval can expose such alternatives, but their literary significance must still be evaluated by a scholar. The appropriate system is therefore not universally one that enforces or disregards theological conventions; its retrieval and evaluation criteria must reflect the hermeneutic objective of the study.

\subsection{Reference Length and Retrieval Performance}

\begin{figure}
\centering
\includegraphics[width=\linewidth]{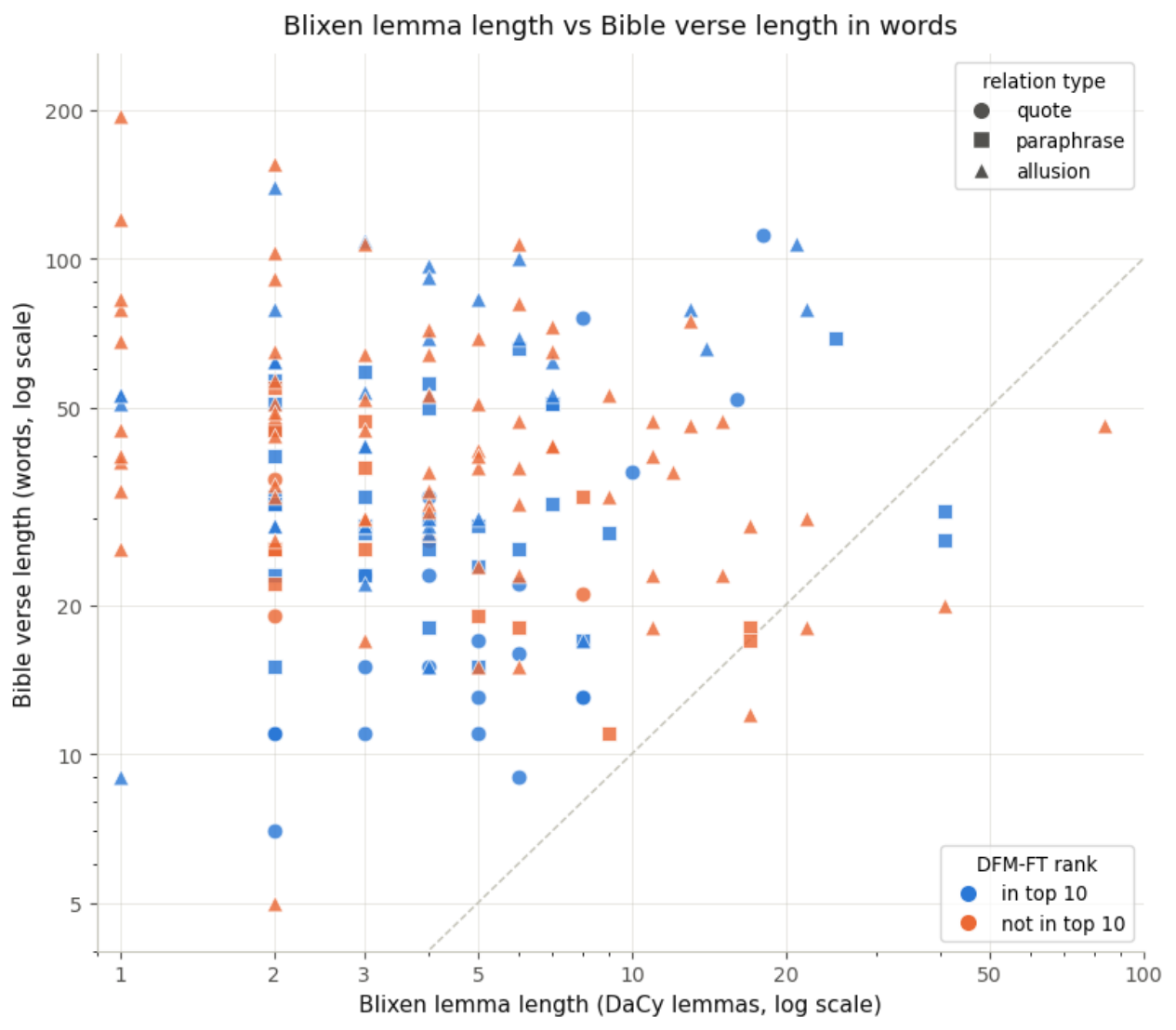}
\caption{Lengths of the Blixen lemmas and their expert-annotated ground-truth biblical passages. Each point represents one reference instance. For a multi-verse reference, passage length is the sum of the word counts of its constituent verses. Shape indicates the automatically assigned relationship type, and colour indicates whether dfm-ft retrieved at least one verse from the ground-truth passage among its ten highest-ranked results (R@10). No clear relationship between text length and retrieval success is apparent.}
\label{fig:reference_length}
\end{figure}

Figure~\ref{fig:reference_length} examines whether retrieval success varies with the lengths of the Blixen lemma and its expert-annotated ground-truth passage. For multi-verse references, the biblical passage length is calculated as the sum of the word counts of all constituent verses. The figure does not show a clear concentration of successful predictions among either shorter or longer lemmas or biblical passages. Nor is there an apparent pattern suggesting that dfm-ft succeeds primarily when the two texts have similar lengths. Within this dataset, retrieval performance therefore does not appear to be systematically associated with reference length.

The successful retrieval of the ten-verse passage in Job shows that the model can overcome a substantial length discrepancy in an individual case. It should not, however, be interpreted as evidence that fine-tuning generally favours long passages or eliminates length-related difficulties.

\section{Conclusion and Future Work}

This study investigated the automatic retrieval of biblical intertexts in Karen Blixen's \textit{Seven Gothic Tales}. Using 189 reference instances derived from a scholarly commentary, we compared sparse and dense retrieval methods and fine-tuned a Danish sentence encoder for the task. The zero-shot results showed that linguistically normalized BM25 remains a strong baseline, particularly for quotations, while dense retrieval offered some advantages for less lexically explicit allusions. Fine-tuning substantially improved dfm-large, increasing its overall R@10 from 0.265 to 0.508 and producing the strongest overall results across the reported metrics. The improvement was most pronounced for paraphrases and allusions, suggesting that task-specific training is especially valuable when direct lexical overlap is limited.

The study also demonstrates a central difficulty in evaluating computational intertextuality: scholarly annotations provide an operational ground truth, but not necessarily an exhaustive inventory of relevant textual relationships. In the expert reassessment, seven of 30 selected dfm-ft predictions originally counted as errors were judged to be meaningful additional references. Retrieval models can therefore support close reading not only by reproducing established annotations, but also by proposing previously unrecorded connections for scholarly evaluation. Such suggestions should be treated as interpretive candidates rather than automatic discoveries, since biblical self-reference and competing citation conventions make relevance dependent partly on the aims of the analysis.

Future work should explore additional fine-tuning approaches, including similarity-based objectives that directly train the model to distinguish degrees of semantic and intertextual relatedness. A larger expert-annotated dataset would make it possible to compare these approaches more reliably and to evaluate automatically derived relationship types against literary judgements. The apparent tension between Old and New Testament matches also warrants systematic investigation. In particular, future experiments could measure whether different retrievers exhibit a bias toward either testament and determine whether fine-tuning reinforces, reduces, or redirects such tendencies.

The retriever could also form the basis of a retrieval-augmented generation system that proposes candidate passages together with evidence-grounded explanations of their possible relationship to the literary text. Such a system could support an iterative workflow in which scholars inspect, reject, or refine suggested intertexts. More broadly, intertextual retrieval offers a demanding benchmark for natural language understanding: success may require recognizing lexical correspondences, paraphrases, recurring motifs, textual transmission, and culturally situated interpretive conventions. Benchmarks developed with literary scholars could therefore serve both computational evaluation and close reading, while also testing forms of language understanding that are poorly captured by standard semantic-similarity tasks. The model thus functions as a heuristic co-reader that augments rather than replaces human judgment and leaves the final interpretive authority to the reader.




\printbibliography




\end{document}